\documentclass{article}
\usepackage{spconf,amsmath,graphicx,amsfonts,amsthm,amssymb,caption, algorithm2e, mathtools, cite, multirow}

\DeclareMathOperator*{\argmin}{arg\,min}

\DeclareMathOperator*{\Ph}{\mathrm{Ph}}

\title{Alternating Phase Langevin Sampling \\with Implicit Denoiser Priors for Phase Retrieval}
\name{Rohun Agrawal$^{\dagger}$\thanks{$^\dagger$This work was supported by the Rita A. and Øistein Skjellum SURF Fellowship}, Oscar Leong}

\address{Computing and Mathematical Sciences, California Institute of Technology}

\begin{document}
%
\maketitle
\begin{abstract}
Phase retrieval is the nonlinear inverse problem of recovering a true signal from its Fourier magnitude measurements. It arises in many applications such as astronomical imaging, X-Ray crystallography, microscopy, and more. The problem is highly ill-posed due to the phase-induced ambiguities and the large number of possible images that can fit to the given measurements. Thus, there's a rich history of enforcing structural priors to improve solutions including sparsity priors and deep-learning-based generative models. However, such priors are often limited in their representational capacity or generalizability to slightly different distributions. Recent advancements in using denoisers as regularizers for non-convex optimization algorithms have shown promising performance and generalization. We present a way of leveraging the prior implicitly learned by a denoiser to solve phase retrieval problems by incorporating it in a classical alternating minimization framework. Compared to performant denoising-based algorithms for phase retrieval, we showcase competitive performance with Fourier measurements on in-distribution images and notable improvement on out-of-distribution images. 
\end{abstract}
\section{Introduction}
\label{sec:intro}

Phase retrieval is the reconstruction problem of obtaining lost phase information of a signal given only its Fourier magnitude. It arises in many physical measurement applications including X-ray crystallography, transmission electron microscopy, astronomical imaging, and more.  
More formally, the phase retrieval problem asks to recover a signal $x^*\in\mathbb{R}^n$ from phaseless measurements of the form
\begin{equation}
b = |Ax^*| + w,
\end{equation}
where $A \in \mathbb{C}^{m \times n}$ is the measurement matrix and $w$ is noise. In the Fourier case, $A$ is replaced with $\mathcal{F}$, the Fourier transform, that operates on $x^*$. 

Due to the phase ambiguity and noise present in the problem, phase retrieval is an ill-posed problem. There are many possible images that can fit to the given measurements, most of them lacking natural structure. Image priors combat this by constraining the space of possible solutions to images that do contain the structure that the object of interest is believed to obey.  There is a rich history of image priors across computational imaging, and we highlight previous priors utilized in phase retrieval.

A common structural prior traditionally enforced in imaging applications is sparsity with respect to a wavelet basis. Several works have considered the use of sparsity priors in phase retrieval to improve reconstruction and yield more efficient algorithms \cite{SparsePRoverview}. However, sparsity priors are limiting as they inherently enforce a linear signal model, which may potentially lack the richness to model more complicated data of interest. On the learning side, there has been tremendous progress on priors in the form of generative models
given by a deep neural network. For example, the works \cite{hand2018phase, ShamshadAhmed21} have shown that generative networks can potentially better represent natural signals compared to sparsity priors and also demonstrate strong recovery performance. One drawback of this approach, however, is that generative models require an abundance of clean data to be trained. Moreover, such generators are often tailored specifically to a data distribution of interest, and struggle to reconstruct out-of-distribution images. These points ultimately limit their use in practical applications. 

An interesting class of (implicit) priors are in the form of neural network-based denoisers, which are models that are trained remove additive noise from images. Denoisers implicitly learn natural image structure through many paired examples of noisy and clean images. Such models have shown impressive performance as regularizers in several linear inverse problems \cite{venkatakrishnan_plug-and-play_2013}. Additionally, these models have shown strong recovery performance as regularizers in phase retrieval and showcase the ability to recover images outside of their training distribution  \cite{metzler_prdeep_2018}. Recently, the work of \cite{kadkhodaie_solving_2021} showed that it was possible to sample from the prior implicit in a denoiser, and demonstrated strong performance in using this prior to help solve linear inverse problems.

\textbf{Our contributions:} In this work, we propose an algorithm called Alternating Phase Langevin Sampling (APLS) to solve phase retrieval problems by combining classical alternating projection methods with modern denoiser priors. Our algorithm incorporates a posterior sampling method to access the implicit prior of a denoiser in an alternating minimization framework. Then, we empirically study our approach in the challenging Fourier measurement regime and showcase competitive results for in-distribution images and significantly outperform other approaches on out-of-distribution images. Our work showcases robustness and generalization performance for such priors, and highlights the potential for incorporating implicit denoiser priors more broadly for nonlinear inverse problems.

\section{Related Work}
\label{sec:background}

Phase retrieval methods can be classified as either convex or non-convex approaches. Convex techniques include semidefinite programming methods, such as PhaseLift \cite{candes2013phaselift}, and methods based on linear programming, such as PhaseMax \cite{Phasemax}. Non-convex methods for phase retrieval generally fall under the alternating projections framework, originating from the Gerchberg-Saxton  \cite{GerchSaxAlg} and Fienup \cite{Fienup87} algorithms, or direct first-order methods, such as Wirtinger Flow \cite{Candes2015}. We now detail some of the alternating minimization approaches that our work is inspired by.

\subsection{Alternating Minimization}
Alternating minimization provides an optimization framework for solving nonlinear inverse problems \cite{netrapalli_phase_2015}. In particular, the nonlinear system $b = |Ax^*|$ can be reduced to a linear system if we had access to the true phase $p$ of $Ax^*$. However, $p$ is not actually known and so the solution to the linear system is obtained by solving
\begin{equation}
\argmin_{|p|=1, x \in \mathbb{C}^n} \|Ax - p \odot b\|_2. 
\end{equation}
Alternating minimization then alternates between updating the phase $p$ and the solution $x$ to iteratively minimize (2). Denoting the phase of $z \in \mathbb{C}$ by $\mathrm{Ph}(z) \vcentcolon= \frac{z}{|z|}$, the phase $p_t$ and signal $x_t$ are updated via the following iteration for each $t \geqslant 1$:
\begin{align*}
    p_{t} & \gets \mathrm{Ph}(Ax_{t-1}), \\
    x_{t } & \gets \argmin_{x\in \mathbb{C}^n}\|Ax -p_{t}\odot b\|_2.
\end{align*} The authors of \cite{netrapalli_phase_2015} showcase geometric convergence of the algorithm with proper initialization and establish sample complexity guarantees for Gaussian measurements. 


\subsection{Denoisers for Regularization}
Denoisers were first used as regularizers in solving inverse problems by \cite{danielyan2017spatially}. This method is termed plug-and-play regularization because a denoiser can be ``plugged in'' as a regularizer when solving an inverse problem to produce more accurate recoveries \cite{venkatakrishnan_plug-and-play_2013}. 

The current state-of-the-art method for solving phase retrieval using a denoiser as a prior is prDeep \cite{metzler_prdeep_2018}. They use the Regularization by Denoising framework (RED) \cite{romano2017little} that explicitly defines the regularizer as
\begin{equation}
R(x) = \frac{\lambda}{2}x^T(x - D(x)),
\label{prdeep-regularizer}
\end{equation}
where $D(x)$ is the denoiser. For the denoiser, a powerful neural network-based denoiser DnCNN \cite{zhang2017beyond} is used. 

\subsection{Prior Implicit in a Denoiser}

Why are denoisers such strong priors for inverse problems? When training to remove additive noise, denoisers must develop an understanding of what the important properties of structured, natural images are. This implicit prior embedded within denoisers was exposed by \cite{kadkhodaie_solving_2021} through Empirical Bayes estimation and was used to achieve competitive performance in solving linear inverse problems.

Specifically, it was shown in \cite{reehorst_regularization_2019} that there is an interesting connection \cite{robbins1992empirical} between the score of the density $p(y)$ induced by images contaminated with Gaussian noise $y = x+z$ with variance $\sigma^2$ and MMSE denoisers: \begin{equation}
   \nabla_y\log p(y) = \frac{\hat{x}(y) - y}{\sigma^2}
\end{equation} where $\hat{x}(y) = \mathbb{E}[x |y]$ is the MMSE estimate, which can be approximated by a denoiser trained on a large paired dataset of clean and noisy images. Based on this observation, \cite{kadkhodaie_solving_2021} proposed an iterative algorithm that samples from a denoiser to take a step in the direction specified by the observation density gradient and gradually reduce the effective noise $\sigma$ to eventually converge to a high-probability region of $p(x)$.
To achieve this, \cite{kadkhodaie_solving_2021} employed a Langevin sampling algorithm derived from Langevin Dynamics to inject a decreasing amount of noise into the gradient ascent procedure.

Since the effective noise is decreasing, the denoiser must be able to adjust to the amount of noise as the stochasticity of the gradient changes. Most denoisers such as the one used in prDeep, are trained to only handle a specific noise value. Instead, the algorithm of \cite{kadkhodaie_solving_2021} uses a bias-free CNN denoiser. By removing all the additive bias terms of the convolutional layers, \cite{mohan_robust_2020} discovered that the resulting denoiser would become robust to the amount of additive noise.

Finally, this Langevin sampling algorithm was used to solve linear inverse problems such as deblurring, super-resolution, and inpainting. Updating the gradient step to handle the conditional density $p(x|Ax=Ax^*)$ allows the algorithm to sample the denoiser in a way that also fits to linear measurements $Ax = Ax^*$. 

\begin{figure*}[htb]\centering
\begin{center}
\captionof{table}{Average PSNR of 128 $\times$ 128 images from Fourier measurements.}
\begin{tabular}{ |l|l|l|l|l|l|l|l|l|l| }
\hline
& \multicolumn{3}{ |c| }{$\alpha = 2$} & \multicolumn{3}{ |c| }{$\alpha = 3$} & \multicolumn{3}{ |c| }{$\alpha = 4$}\\
\hline
 & HIO & prDeep & APLS & HIO & prDeep & APLS & HIO & prDeep & APLS\\ \hline
 Natural & 18.72 & \textbf{30.02} & 29.34 & 19.04 & 28.62 & \textbf{28.69} & 18.56 & \textbf{28.05} & 27.67 \\ \hline
 Unnatural & 19.75 & 21.9 & \textbf{22.32} & 21.24 & 24.08 & \textbf{25.5} & 18.57 & 22.29 & \textbf{22.44} \\

\hline
\end{tabular}
\end{center}
\end{figure*}
\begin{figure*}[htb]\centering
\begin{tabular}{cccc}
\includegraphics[width=0.22\textwidth]{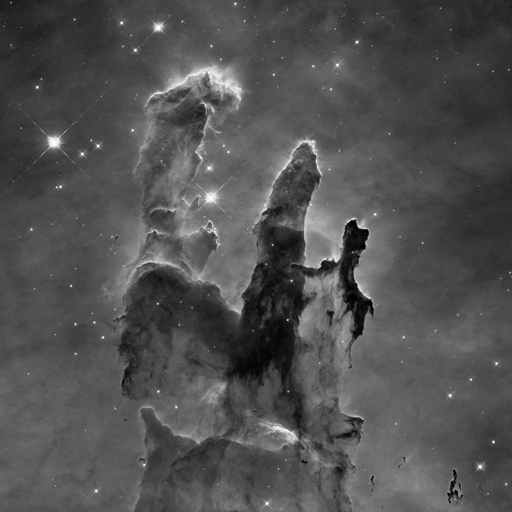} &
\includegraphics[width=0.22\textwidth]{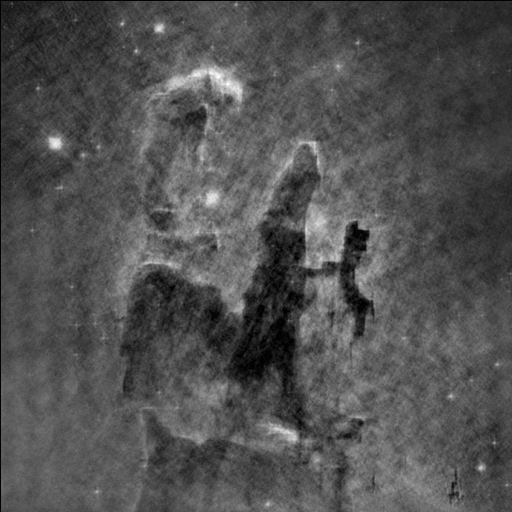} &
\includegraphics[width=0.22\textwidth]{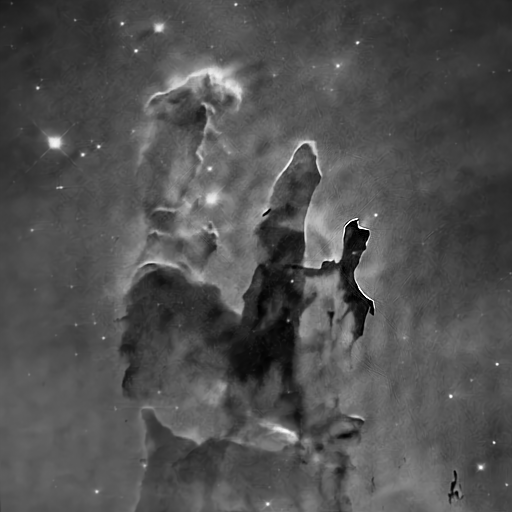} &
\includegraphics[width=0.22\textwidth]{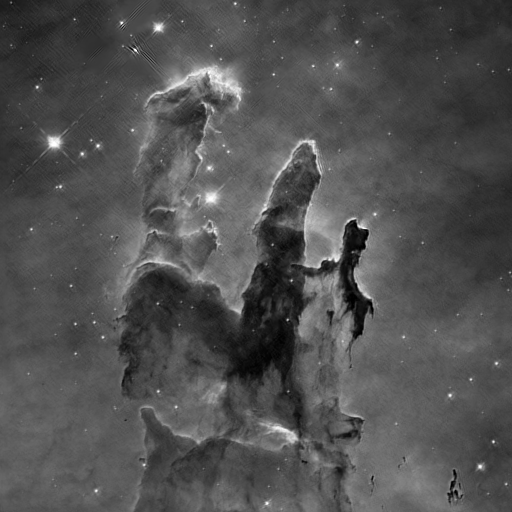}
\\
Original  & HIO (25.59) & prDeep (28.82) & \textbf{APLS (35.37)}  \\[6pt]
\end{tabular}
\caption*{Figure 1: Reconstructions (PSNR) of a 512 $\times$ 512 unnatural image from Fourier measurements.}

\end{figure*}
\section{Approach}

We aim to incorporate an implicit denoiser prior into an efficient algorithm to solve phase retrieval problems. The denoiser prior has learned universal properties of natural images, and, thus, could potentially provide strong generalization guarantees to different datasets. One issue, however, is that the approach of \cite{kadkhodaie_solving_2021} only allowed for sampling from the posterior conditioned on \textit{linear} measurements $p(x | Ax = Ax^*)$, rendering an extension to nonlinear inverse problems challenging. Inspired by alternating projection approaches, we propose an alternating minimization-sampling procedure that 1) fixes a candidate phase $p_t$ and then 2) updates the signal by sampling from the posterior $x_t \sim p(x | Ax = p_t \odot b)$. We first describe the Langevin sampling algorithm for sampling from the posterior of the image conditioned on the linear part of the measurements.

\subsection{Langevin Sampling}

As noted in \cite{kadkhodaie_solving_2021}, the residual of the denoiser $D(x) := \hat{x}(y) - y$ is proportional to the gradient of log observation density. This result can be utilized to perform the following update steps to approximately sample from the conditional density $p(x | Ax = x^c)$, as outlined in Algorithm 1. The main update step is
\begin{equation}
x \gets x + hd_x + \gamma z
\end{equation}
where $hd_x$ is the gradient step with step size $h$ and $\gamma z$ is injected noise. The quantity $\gamma$ controls the proportion of injected noise $z \sim \mathcal{N}(0, I)$ and is given by
\begin{equation}
\gamma^2 \gets ((1 - \beta h)^2 - (1-h)^2)\sigma^2,
\end{equation}
where $\beta \in (0, 1]$ is a hyperparameter. $\sigma^2 = \frac{\| d \|_2^2}{N}$ ($N$ is the size of the image) adjusts the amount of injected noise in accordance to the size of the gradient step. The gradient step, $d_x$, is
\begin{equation}
d_x \gets (I - A^TA)D(x) + A^T(x^c - Ax),
\end{equation}
and uses $D$, the residual of the denoiser, to approximately sample from the conditional distribution $p(x|Ax=x^c)$. Note that the quantity $h_t = \frac{h_0t}{1 + h_0(t-1)}$ acts as an accelerator for the algorithm by increasing the step size and decreasing the amount of injected noise over time. In our case, since $A$ is the discrete Fourier transform, it satisfies assumptions of row orthogonality made by \cite{kadkhodaie_solving_2021} for Algorithm 1. 

\label{sec:algorithm}
\RestyleAlgo{ruled}
\begin{algorithm}
\caption{Posterior Sampling via the Implicit Denoiser Prior}
\KwData{$x^c$, $A$, $T_1$, $x_0$, $D$, $h_0$ $\beta$}
\KwResult{$x_{T_1}$}
\For {$t = 1, \dots, T_1$}{
$h_t \gets \frac{h_0t}{1 + h_0(t-1)}$

$d_t \gets (I - A^TA)D(x_{t-1}) + A^T(x^c - Ax_{t-1})$

$\sigma^2_t \gets \frac{\lVert d_t \rVert ^2}{N}$

$\gamma^2_t \gets ((1 - \beta h_t)^2 - (1-h_t)^2)\sigma^2_t$

$z_t \sim \mathcal{N}(0, I)$

$x_t \gets x_{t-1} + h_td_t + \gamma_t z_t$
}

\end{algorithm}

\begin{figure}
\begin{minipage}[b]{1.0\linewidth}
  \centering

  \centerline{\includegraphics[width=8cm]{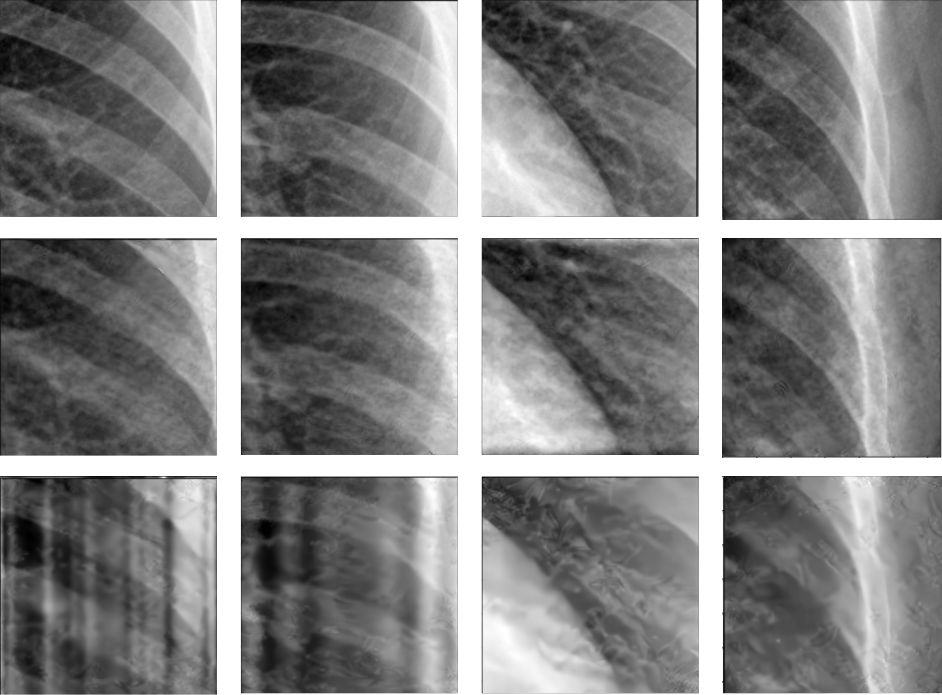}}
  Figure 2: Examples of Chest X-Ray image reconstructions from Fourier measurements. First Row: Original. Second Row: APLS (Ours). Third Row: prDeep
\end{minipage}
\end{figure}

\subsection{APLS: Alternating Phase Langevin Sampling}



\begin{algorithm}
\caption{APLS}
\KwData{$b$, $A$, $T_1$, $T_2$, $x_0$, $D$, $h_0$, $\beta$}
\KwResult{$x_{T_2}$}
\For {$t = 1, \dots, T_2$}{
    
    $p_t \gets $ Ph$(Ax_{t-1})$
    
    $x_{t} \gets $Algorithm 1 with $\{x^c = p_t \odot b, A, T_1, x_0 = x_t, D,h_0, \beta\}$
}
\end{algorithm}

We use the Alternating Minimization framework to decouple the non-linearity of the phase retrieval problem to incorporate a denoiser prior with Langevin Sampling. Specifically, we perform an alternating minimization-sampling algorithm that first fixes the phase and then approximately samples from the posterior conditioned on the phase-induced measurements by utilizing the implicit denoiser prior. That is, for initial points $p_0$ and $x_0$, we compute for $t \geqslant 1$, \begin{equation}
    p_{t}  = \Ph(Ax_{t-1}), \\
    x_{t}  \sim p(x | Ax = p_t \odot b). 
\end{equation}  Additionally, we enforce the gradient of the Langevin Sampling to account for the phase $p_t$ by providing it the measurements $x^c = p_t \odot b$. 
We outline more details in Algorithm 2.

\section{Experimental Results}
\label{sec:experimentalresults}


The classical phase retrieval problem is to recover true signal $x^*$ from measurements $b = |\mathcal{F}(x^*)| + w$. Poisson shot noise is a common noise model that appears in many phase retrieval applications \cite{yeh2015experimental}. We approximate this as
\begin{equation}
b^2 = |\mathcal{F}(x^*)|^2 + w, w \sim \mathcal{N}(0, \alpha^2\text{Diag}(|\mathcal{F}(x^*)|^2))
\end{equation}
where $\alpha$ is a parameter that increases with the noise level. We use 4$\times$ over-sampled Fourier measurements by placing an $n \times n$ image in the center of a $2n \times 2n$ square and taking its 2D Fourier transform. The location of the image in this square is the support which we assume is known a priori. 

As in \cite{metzler_prdeep_2018}, for the initial input $x_0$ we use the output of the Hybrid Input-Output (HIO) algorithm \cite{Fienup87}. We take 50 random initializations $x_1, x_2, \dots x_{50}$ and run HIO for 50 iterations. We take the $x_i$ with the lowest residual $\| b - |\mathcal{F}(x_i)| \|_2$ and run HIO for another 1000 iterations with the initial iterate $x_i$. The result is used as the input $x_0$ for APLS. Our algorithm is initialized with $h_0 = 0.1$, $\beta = 0.0001$, and $T_2 = 500$. Empirically, we achieved best results when the number of iterations to run the Langevin Sampling algorithm between phase updates was just $T_1 = 1$. 

We compare to prDeep which is the state-of-the-art using the regularization by denoising framework. We use 4 DnCNN denoisers trained on the Berkeley Segmentation Dataset with noise standard deviations of 50, 25, 15, and 5 to implement the pipeline used in the original paper.

We evaluate the reconstructions of HIO, prDeep, and our algorithm on a set of 6 natural grayscale images and a set 6 unnatural grayscale images. These images are widely used for evaluating inverse problem reconstruction quality. We calculate the Peak Signal to Noise Ratio, $\text{PSNR} = 10\log_{10}(\frac{255^2}{\text{mean}((\hat{x} - x^*))})$, to assess the reconstruction quality between reconstructed image $\hat{x}$ and original image $x^*$. Table 1 displays the average PSNR across both test sets. We see that our method is comparable on natural images and we showcase considerable improvement for unnatural images over all noise levels. Finally, we compared the performance of prDeep to APLS on the Shenzhen chest x-ray test set ($60$ images) \cite{Stefan_Jaeger_2014}. At $\alpha = 3$, prDeep achieved an average PSNR of 33.68 while APLS achieved better with an average PSNR of 35.04. Figure 2 shows that APLS displays far less artifacts in its reconstructions. 

\section{Conclusion}
\label{sec:conclusion}

We proposed the Alternating Phase Langevin Sampling (APLS) algorithm for phase retrieval, which incorporates an implicit denoiser prior in an alternating minimization framework. We demonstrated competitive performance with prDeep, a strong baseline incorporating a denoiser as a regularizer for phase retrieval. For natural images, our performance was comparable while for unnatural, out-of-distribution images, we showcased superior performance. This work showcases the benefits of extracting the prior in a denoiser more directly. There are many interesting avenues of future work, including applying our approach to phase retrieval problems in more scientific domains, such as astronomical imaging, and theoretically analyzing the convergence and sample complexity of our method.


\label{sec:refs}
\bibliographystyle{IEEEbib}
\bibliography{references_zotero, dpr, refs}

\end{document}